%% file: main.tex
\documentclass[final]{cvpr}

\usepackage{times}
\usepackage{epsfig}
\usepackage{graphicx}
\usepackage{amsmath}
\usepackage{amssymb}
\usepackage{color}
\usepackage{xcolor}
\usepackage{booktabs}
\usepackage{tabularx}

\definecolor{turquoise}{cmyk}{0.65,0,0.1,0.3}
\definecolor{purple}{rgb}{0.65,0,0.65}
\definecolor{dark_green}{rgb}{0, 0.5, 0}
\definecolor{orange}{rgb}{0.8, 0.6, 0.2}
\definecolor{red}{rgb}{0.8, 0.2, 0.2}
\definecolor{darkred}{rgb}{0.6, 0.1, 0.05}
\definecolor{blueish}{rgb}{0.0, 0.3, .6}
\definecolor{light_gray}{rgb}{0.7, 0.7, .7}
\definecolor{pink}{rgb}{1, 0, 1}
\definecolor{greyblue}{rgb}{0.25, 0.25, 1}
\definecolor{tab_blue}{HTML}{1f77b4}
\definecolor{tab_orange}{HTML}{ff7f0e}
\definecolor{LightRed}{rgb}{0.99,0.89,0.89}
\definecolor{mesh_misty_rose}{HTML}{e6aaa3}
\definecolor{mesh_yellow}{HTML}{ffba00}
\definecolor{MyDarkBlue}{rgb}{0.02,0.02,0.6}
\definecolor{gold}{rgb}{0.7, 0.5, 0}

\usepackage[pagebackref,breaklinks,colorlinks]{hyperref}

\def\confName{CVPR}
\def\confYear{2024}
\graphicspath{{./figs/}{./figs/result/}}

\input{definitions}

\begin{document}

\title{\methodname: Parametric Control of Material Properties with Diffusion Models}

\author{Prafull Sharma\textsuperscript{*,1,2} \quad Varun Jampani\textsuperscript{$\dagger$,2} \quad Yuanzhen Li\textsuperscript{2} \quad Dmitry Lagun\textsuperscript{2} \\ Fredo Durand\textsuperscript{2} \quad Bill Freeman\textsuperscript{1,2} \quad Mark Matthews\textsuperscript{2} \\
${^1}$MIT CSAIL \quad $^{2}$Google Research \\
\href{http://www.prafullsharma.net/alchemist}{\texttt{www.prafullsharma.net/alchemist}}
}

\vspace{-3em}

\twocolumn[{
    \renewcommand\twocolumn[1][]{#1}
    \maketitle
    \begin{center}
    \centering
    \input{figures/tex/teaser}

    \end{center}
}]

\begin{abstract}
\vspace{-1em}

\input{sections/0_abstract}
\end{abstract}

\input{sections/1_introduction}

\input{sections/2_related_work}

\input{sections/3_method}
\input{sections/4_results}

\input{sections/5_discussion}

\input{sections/6_conclusion}

\section{Acknowledgements}
We would like to thank Forrester Cole, Charles Herrmann, Junhwa Hur, and Nataniel Ruiz for helpful discussions. Thanks to Shriya Kumar and Parimarjan Negi for proofreading the submission.

{\small
\bibliographystyle{ieee_fullname}
\bibliography{main}
}

\end{document}

%% file: definitions.tex
\newcommand{\methodname}{Alchemist\xspace}

\newcommand{\pplus}{$\mathcal{P}+$}

\newcommand{\strength}{s}
\newcommand{\strengths}{\mathbf{s}}
\newcommand{\albedo}{s_{a}}
\newcommand{\roughness}{s_{r}}
\newcommand{\metallic}{s_{m}}

\newcommand{\image}{\mathbf{I}}
\newcommand{\loss}{\mathcal{L}}
\newcommand{\contextimage}{I_c}
\newcommand{\editedimage}{I_e}
\newcommand{\encoder}{\mathcal{E}}
\newcommand{\latent}{z}
\newcommand{\decoder}{\mathcal{D}}
\newcommand{\noise}{\epsilon}
\newcommand{\dtime}{t}
\newcommand{\prompt}{p}
\newcommand{\denoise}{\noise_\theta}
\newcommand{\nullweight}{w_{null}}
\newcommand{\nullthreshold}{\tau}

\newcommand{\Fig}[1]{Fig.~\ref{fig:#1}}
\newcommand{\Figure}[1]{Figure~\ref{fig:#1}}

\newcommand{\Table}[1]{Table~\ref{tab:#1}}

%% file: figures/tex/teaser.tex
\captionsetup{type=figure}
\vspace{-2em}
\includegraphics[width=\linewidth]{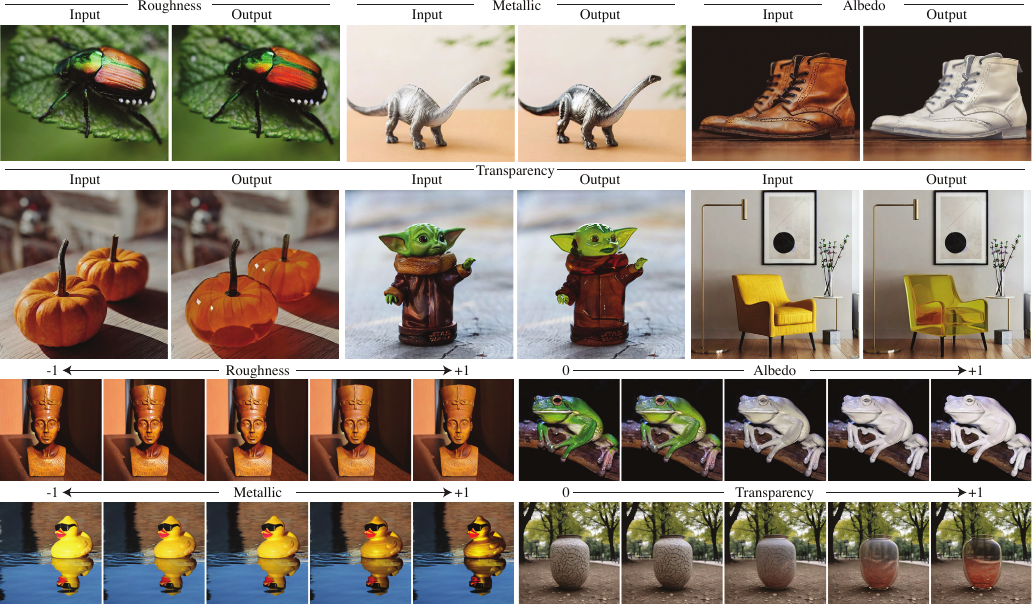}
\captionof{figure}{
    \textbf{Overview.} Our method, \methodname, edits material properties of objects in input images by relative attribute strength $\strength$.
    \textbf{Top:} We set the strength $\strength=1$, resulting in a beetle without specular hightlights, a dark metallic dinosaur, and boot with gray albedo. Our model generates plausible transparency including the light tint, caustics, and hallucinated plausible details behind the object. \textbf{Bottom:} We demonstrate smooth edits for linearly chosen strength values.
}
\label{fig:teaser}

%% file: sections/0_abstract.tex
We propose a method to control material attributes of objects like roughness, metallic, albedo, and transparency in real images.\let\thefootnote\relax\footnote{*This research was performed while Prafull Sharma was at Google.\\$~~~~~~~~^{\dagger}$Varun Jampani is now at Stability AI.}
Our method capitalizes on the generative prior of text-to-image models known for photorealism, employing a scalar value and instructions to alter low-level material properties.
Addressing the lack of datasets with controlled material attributes, we generated an object-centric synthetic dataset with physically-based materials.
Fine-tuning a modified pre-trained text-to-image model on this synthetic dataset enables us to edit material properties in real-world images while preserving all other attributes. 
We show the potential application of our model to material edited NeRFs.

%% file: sections/1_introduction.tex
\vspace{-2em}
\section{Introduction}
Achieving fine-grained control over material properties of objects in images is a complex task with wide commercial applications beyond computer graphics. This ability is particularly relevant in image editing, advertising, and image forensics. We propose a method for precise editing of material properties in images, harnessing the photorealistic generative prior of text-to-image models. We specifically target four key material properties: roughness, metallic, albedo, and transparency. 
Our results illustrate that generative text-to-image models contain a strong understanding of light transport which can be leveraged for precise control of these material properties. 
The physics of light transport affects the appearance of the object. How we view the objects is an interplay of physical factors such as surface geometry, illumination sources, camera intrinsics, color science, sensor linearity, and tone-mapping. However, the most significant of these factors is material properties.

In computer graphics, Bidirectional Reflectance Distribution Functions (BRDFs)~\cite{cook1982reflectance,he1992fast,he1991comprehensive} define material properties which led to the development of principled and physically based BRDF models~\cite{Burley2012pbs}. Prior methods typically employed an inverse rendering approach to disentangle and estimate complex scene attributes like geometry and illumination for material modification~\cite{khan2006image}. Recent work by Subias et al. proposed a GAN-based method trained on synthetic data for perceptual material edits, focusing on metallic and roughness parameters, necessitating the masking of the targeted real-world object~\cite{subias2023wild}. Our approach uses the generative prior of text-to-image models. We directly modify real-world images in pixel space, eliminating the need for auxiliary information such as explicit 3D geometry or depth maps, environment maps, and material annotations, thereby bypassing the process of accurately estimating object and scene-level properties.

Manipulating material properties in pixel space using a pre-trained text-to-image model presents two main challenges.
First, the scarcity of real-world datasets with precisely labeled material properties makes generalizing from supervised training difficult. Second, text-to-image models are trained with textual descriptions like "gold," "wood," or "plastic," which often lack fine-grained details about the material. This issue is compounded by the inherent disconnect between the discrete nature of words and the continuous nature of material parameters.

To overcome the first challenge, we render a synthetic dataset featuring physically-based materials and environment maps, thus addressing the need for fine-grained annotations of material properties. For the second challenge, we employ extra input channels to an off-the-shelf diffusion model, refining this model with an instruction-based process inspired by InstructPix2Pix~\cite{brooks2023instructpix2pix}. Despite being trained on only 500 synthetic scenes comprising 100 unique 3D objects, our model effectively generalizes the control of material properties to real input images, offering a solution to the issue of continuous control. 

To summarize, we present a method that utilizes a pre-trained text-to-image model to manipulate fine-grained material properties in images. Our approach offers an alternative to traditional rendering pipelines, eliminating the need for detailed auxiliary information. The key contributions of our method are as follows:
\begin{enumerate}
    \setlength{\itemsep}{0pt}
    \item We introduce an image-to-image diffusion model for parametric control of low-level material properties, demonstrating smooth edits of roughness, metallic, albedo and transparency. 
    \item We render a synthetic dataset of fine-grained material edits using 100 3D objects and randomized environment maps, cameras, and base materials.
    \item Our proposed model generalizes to real images despite being trained on synthetic data. 
\end{enumerate}

%% file: sections/2_related_work.tex
\section{Related Work}

\paragraph{Diffusion models for image generation.} Denoising Diffusion Probabalistic Models (DDPMs) have been an active focus of the research community~\cite{dhariwal2021diffusion,ho2020denoising,ho2022cascaded,ho2022classifier,song2019generative,karras2022elucidating,kang2023scaling,crowson2022vqgan} for their excellent photorealistic image generation capabilities from text prompts~\cite{saharia2022photorealistic,ramesh2022hierarchical,rombach2021high,nichol2021glide}. Image-to-image tasks are possible by modifying the denoising network to accept image inputs, allowing style-transfer~\cite{sohn2023styledrop}, inpainting, uncropping, super-resolution, and JPEG compression~\cite{saharia2022palette}. Furthermore, the generative priors of 2D diffusion models have been utilized towards novel-view synthesis, 3D generation, and stylistic 3D editing~\cite{sjc,syncdreamer,shi2023mvdream,ssdnerf, raj2023dreambooth3d,haque2023instruct,sella2023voxe,zhuang2023dreameditor,Yu2023EditDiffNeRF,Liu2023Zero1to3ZO,Fridman2023SceneScapeTC,Hllein2023Text2RoomET,Tsalicoglou2023TextMeshGO,seo2023let,chen2023it3d,Xu2023InstructP2P,poole2022dreamfusion}. Our image-to-image method leverages and further controls this learned prior of DDPMs.

\vspace{-1em}
\paragraph{Control in generative models.} Controlling generative model output remains an active area of study with many works proposing text-based methods~\cite{kim2022diffusionclip,voynov2023p+,hertz2022prompt,Patashnik_2021_ICCV,kawar2022imagic,tao2022net,brooks2023instructpix2pix,liu2020open,avrahami2022blended,cong2023attribute,ge2023expressive,tumanyan2023plug,mokady2023null,cao2023masactrl}. 
Other works proposed alternative control inputs such as depth maps, sketches~\cite{ye2023ip,voynov2023sketch}, paint strokes~\cite{meng2021sdedit},  identity~\cite{xiao2023fastcomposer,ma2023subject}, or photo collections~\cite{ruiz2022dreambooth,ruiz2023hyperdreambooth,shi2023instantbooth,kumari2023multi}.
Prompt-to-Prompt~\cite{hertz2022prompt}, \pplus~\cite{voynov2023p+}, and Null-text inversion~\cite{mokady2023null} present editing techniques based on reweighting of cross-attention maps. ControlNet~\cite{zhang2023adding} and T2I-Adapter~\cite{mou2023t2i} demonstrate control through spatial inputs defining mid-level information. Generated images from diffusion models can also incorporate new subjects from an image collection using a small number of exemplars~\cite{ruiz2022dreambooth,gal2022image,chen2023subject,ruiz2023hyperdreambooth,shi2023instantbooth,wei2023elite,kumari2023multi}. While these works control high and mid-level information about objects, control of low-level properties such as materials remains a challenge for them, leading us to our present line of study.

\vspace{-1em}
\paragraph{Material understanding and editing.}

Editing materials in images is a significant challenge, requiring a strong understanding of image formation. Human vision research has extensively explored how attributes like albedo, roughness, illumination, and geometry affect object perception~\cite{motoyoshi2012variability,nishida1998use,obein2004difference,olkkonen2010perceived,fleming2003real,doerschner2010estimating,marlow2012perception,fleming2014visual,storrs2021unsupervised,fleming2014visual}.

\input{figures/tex/architecture}

Image based material editing was introduced by Khan et al. presenting simple material operations using depth estimates\cite{khan2006image}. Subsequent works demonstrated disentanglement of material and lighting with a statistical prior~\cite{lombardi2012reflectance}, editing gloss appearance~\cite{manabe2021glossy,boyadzhiev2015band}, intrisic image decomposition~\cite{liu2017material}, and 2D editing of material exemplars~\cite{zsolnai2020photorealistic}. We forego these ``decompositional'' approaches and instead leverage the largely self-supervised prior of DDPMs for direct editing in pixel-space.

Generative models, particularly Generative Adversarial Networks (GANs)~\cite{goodfellow2020generative}, have been investigated for their ability to alter material perception, focusing on gloss and metallic properties~\cite{delanoy2022generative,subias2023wild}. The application of semantic and material editing in NeRFs has also been explored using text-prompts and semantic masks~\cite{haque2023instructnerf2nerf,Zhou2023RePaintNeRF}.

%% file: figures/tex/architecture.tex
\begin{figure*}[t]
\centering
\includegraphics[width=1.0\linewidth]{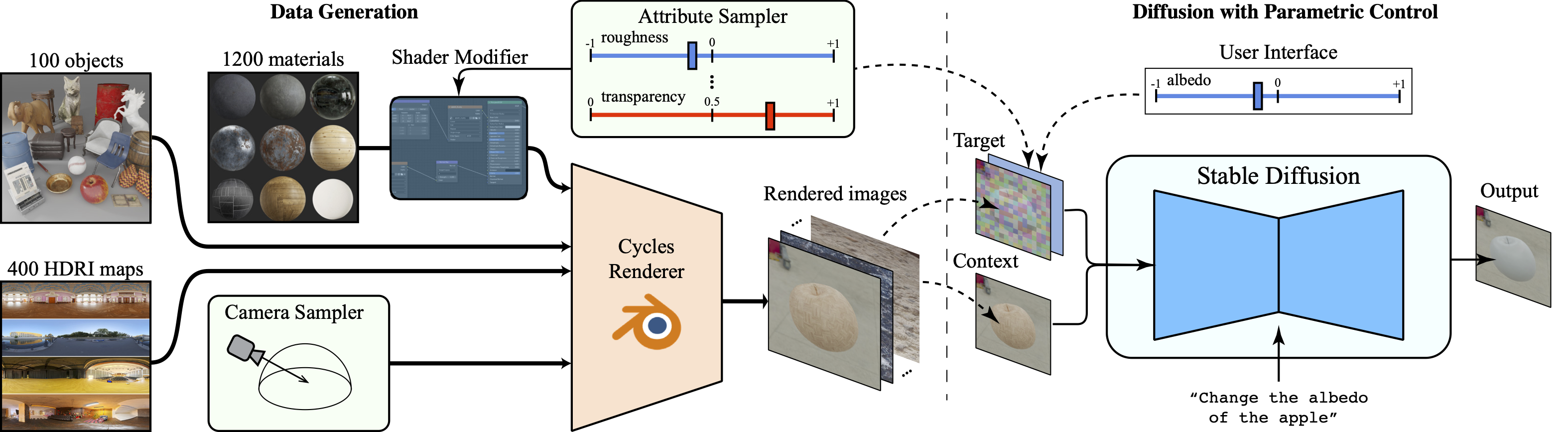}
\caption{\textbf{Method.} We generate a synthetic dataset by taking each of 100 objects, applying randomized materials and illumination maps, and modifying the shading network according to randomly sampled attribute strength $\strength$. Each object is rendered from 15 randomized cameras (see Section~\ref{sec:method} for details). During training we provide the $\strength=0$ image as context and randomly choose a target image of known attribute strength. At test time we provide the user-input context image and edit strength. 
}
\label{fig:architecture}
\vspace{-1em}
\end{figure*}

%% file: sections/3_method.tex
\section{Method}\label{sec:method}
\input{figures/tex/dataset_samples}

There is no existing object-centric dataset that precisely varies only single material attributes. 
Curating such a real world dataset would be infeasible due to the difficulty of creating physical objects in this way with known parameters.
Therefore, we opt to render a synthetic dataset, giving us full control of material attributes. Using this data, we propose a method to perform material attribute control given a context image, instruction, and a scalar value defining the desired relative attribute change. The method is based on latent diffusion model for text-to-image generation with modification that allows us to condition the network on the relative attribute strength. 

\subsection{Datasets}

We render our dataset with the Cycles renderer from Blender~\cite{blender}, using publicly available 3D assets, physically-based materials, and environment maps. 
Each scene begins with one of 100 unique object meshes from polyhaven.com. 
Each of these is paired with five randomly chosen materials of the 1200 available from ambientcg.com, and illuminated with one of 400 environment maps. 
The material is a Principled BRDF shader node, the base shader in Blender.
The base configuration of the material is kept as a control defined as 0 strength change for each of the attributes.
This control serves as the context input image to the method against which relative changes in roughness, metallic, albedo, and transparency are applied, sampling 10 random relative values for each attribute, the details of which are described below.
Finally, we render 15 images of each setup using different camera viewpoints and intrinsics. 
This creates a wide combination of scenes with diversity in material, lighting, and background conditions. 
Samples from the rendered dataset are presented in \Figure{dataset_samples}. 

\vspace{1mm}
\noindent \textbf{Roughness and Metallic.} For both roughness and metallic properties, we operate in an additive framework. In the case when the material has the associated map for roughness or metallic, we use an additive node yielding a parametric control between [-1, 1]. For materials where either of these spatial maps are missing, we control the attribute control directly as a constant map, assuming the base 0.5 as the control state of the attribute. Note that these values are clamped between [0, 1] so in some cases, further increasing or decreasing the roughness does not result in any change in the rendered image. We account for this by under-sampling such images where the gradient of change is constant.

Reducing the roughness value results in a surface that reflects light more uniformly and sharply, giving it a glossy or shiny appearance. On the other hand, increasing the roughness value leads to a more diffused light reflection, making the surface appear matte or dull.
Low metallic value results in appearance predominantly determined by the base color, as in the case of plastic and wood. Increasing the metallic leads to the surface absorbing more of the incoming light, resulting in a darker appearance of the object.

\vspace{1mm}
\noindent \textbf{Albedo.} We implement a color mixing between the original albedo map of the object and a spatially constant gray (RGB = 0.5) albedo map. The parametric controller operates between 0 and 1, where 0 corresponds to the original albedo, and 1 corresponds to completely gray albedo. This can be considered as detexturing the albedo and can be interesting when combined with roughness and metallic parameters to achieve a mirror-like or a shading-only image.

\vspace{1mm}
\noindent \textbf{Transparency.} We introduce the ability to control transparency by controlling the transmission value in the BSDF shader node. The attribute value is chosen to be in range [0, 1]. For a chosen transmission value t, we choose to reduce the roughness and metallic component in an additive manner by t, and also add a white overlay to the albedo to increase the intensity of the appeared color. For the value of 0, we expect the same opaque object and at 1, we would get a transparent version of the object, making it appear as if it was made of glass. Note that we made the choice to retain the effect of the albedo resulting in a fine tint on the object.

\subsection{Parametric Control in Diffusion Models}

The rendered synthetic data is used to finetune an image-to-image diffusion model conditioned on relative attribute strength and a generic text instruction providing parametric control over material properties. We operate in latent space using Stable Diffusion 1.5, a widely adopted text-to-image latent diffusion model. 

Diffusion models perform sequential denoising on noisy input samples, directing them towards the dataset distribution by maximizing a score function~\cite{sohl2015deep}.
A noising process is defined over timesteps $\dtime \in T$, resulting in a normal distribution at $T$.
We operate in latent space by using a pre-trained variational encoder $\encoder$ and ddecoder $\decoder$~\cite{kingma2013auto}, a potent aid to conditional image generation~\cite{rombach2021high}. 
Training draws an image sample $\image$ from the dataset, encodes it into a latent $\latent=\encoder(\image)$, then noises it at $\dtime$ as $\latent_\dtime$.
A denoising network $\denoise$ predicts the added noise given the latent $\latent_\dtime$, diffusion time $\dtime$, and conditioning variables. 

Our image-to-image model is conditioned on an input image to be edited, provided as $\encoder(\contextimage)$ concatenated to the latent being denoised $\latent_\dtime$. 
Text conditioning is provided via cross-attention layers using a generic prompt, $\prompt=$ ``\textit{Change the $\texttt{<attribute\_name>}$ of the $\texttt{<object\_class>}.$}'' Since textual CLIP embeddings~\cite{radford2021learning} do not encode fine-grained information well~\cite{paiss2023teaching}, prompt-only conditioning of $s$ expressed textually (i.e. ``\textit{Change the roughness of the apple by 0.57.}'') yields inconsistent output.
To facilitate relative attribute strength conditioning, we also concatenate a constant scalar grid of edit strength $\strength$.

We initialize the weights of our denoising network with the pre-trained checkpoint of InstructPix2Pix~\cite{brooks2023instructpix2pix}, providing an image editing prior and understanding of instructive prompts. During training (\Fig{architecture}), we minimizes the loss:
\begin{equation}
  \loss = \mathbb{E}_{\encoder(\image), \encoder(\contextimage), \strengths, \prompt, \noise \sim \mathcal{N}(0, 1), \dtime}
  \left[||
    \noise - \noise_\Theta(\latent_\dtime, \dtime, \encoder(\contextimage), \strengths, \prompt)
  ||^2_2\right]
\end{equation}
\input{figures/tex/single_attribute_qual}
\vspace{-1em}

We always provide the $\strength=0$ image as context $\contextimage$ , and draw an edited image $\editedimage$ at random for noising. Since we always render an $\strength=0$ sample, and other $\strength$ are chosen with stratified sampling, our distribution has a slight bias towards zero. Since many edit strengths may have little effect (i.e. we cannot lower the roughness of an object with 0 roughness), we find that providing too many of these examples biases the network towards inaction. We therefore downweight such null examples, defined as $|| \contextimage - \editedimage || ^2 < \nullthreshold$, by $\nullweight$ via rejection sampling. In practice we set $\nullweight = 0.80$, $\nullthreshold=0.05$. We train with $\texttt{fp16}$ precision for 10k steps using Adam~\cite{kingma2014adam} and learning rate of \texttt{5e-5}. We use the text encoders and noise schedule from Stable Diffusion.

At test time we provide a held out image as context $\contextimage$, edit strength $\strength$, and prompt $\prompt$ for the object class of the input image. We denoise for $20$ steps using the DPM-solver$++$ based noise scheduler~\cite{lu2022dpm}.

\vspace{1mm}
\noindent \textbf{Multi-attribute editing.} We edit multiple attributes in a single diffusion pass by concatenating more than one edit strength, drawn from $\{\albedo,\roughness\,\metallic\}$ giving us $[\latent_\dtime | \encoder(\contextimage) | \strengths]$ as the final UNet input, where $|$ is concatenation.

\vspace{1mm}
\noindent \textbf{Classifier-free guidance.} Ho et al.~\cite{ho2022classifier} proposed classifier-free guidance (CGF) to improve visual quality and faithfulness of images generated by diffusion models. We retain the same CFG setup as InstructPix2Pix for both image and prompt conditioning. We do not however impose CFG with respect to the relative attribute strengths $\strengths$. We want the network to be faithful to edit strength and forced to reason about incoming material attributes. As $\strength$ can be 0 by definition of the problem itself, and downweighted as described above, we did not find CFG on $\strengths$ necessary.

We will release the dataset generation pipeline, image renderings with metadata, and the training code.

%% file: figures/tex/dataset_samples.tex
\begin{figure}[t]
    \centering
    \includegraphics[width=\linewidth]{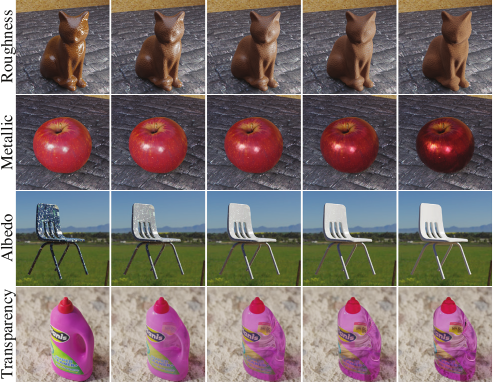}
    \caption{\textbf{Synthetic dataset.} Samples from our synthetic dataset illustrating appearance change for a linear attribute change.
    }
    \label{fig:dataset_samples}
\end{figure}

%% file: figures/tex/single_attribute_qual.tex
\begin{figure*}[t]
    \centering
    \includegraphics[width=\linewidth]{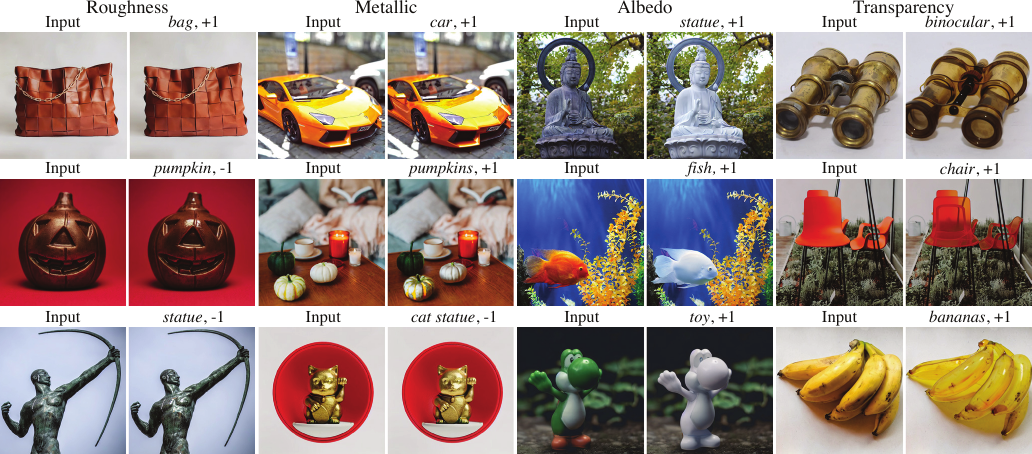}
    
    \caption{\textbf{Single-attribute editing results.} Outputs from our model trained on individual attributes. Left of columns are held-out input and right are model output (\textit{object class}, $\strength$).
    Increased ``Roughness'' replaces specular highlights on the bag with base albedo.
    ``Metallic'' varies contributions from albedo and shine in regions of the pumpkins and cat.
    Surfaces are changed to a flat grey ``Albedo'' revealing object illumination.
    ``Transparency'' preserves object tint while inpainting background and hallucinating plausible hidden structures and caustics.
    }
    \label{fig:single_attribute_qual}
    \vspace{-1em}
\end{figure*}

%% file: sections/4_results.tex
\section{Results}\label{sec:results}
We present qualitative analysis demonstrating the generalization capability of our model to real images despite being trained on synthetic data. Comparisons to baselines show the effectiveness of our model for fine-grained material editing, further supported by a user study. 
We extend the use of our model to NeRF material editing on the DTU dataset~\cite{mildenhall2020nerf,dtu_mvs}.
\subsection{Results on Real Images}
\input{figures/tex/qualitative_results}
We demonstrate the effectiveness of our technique through editing material attributes, one at a time, for real unseen images, in \Figure{single_attribute_qual}. For each of the material attributes we use a separate model trained only on that attribute. We observe that the model outputs preserve geometry and take the global illumination into account.

\vspace{1mm}
\noindent \textbf{Roughness.} As the roughness is increased, the output shows removal of the specular highlights replaced by estimate of the base albedo. The highlights are amplified when the roughness is reduced as shown in the case of the pumpkin and statue. 

\vspace{1mm}
\noindent \textbf{Metallic.} The increase in the metallic component of the car and pumpkin results in dampening of the base albedo and increase in the shine on the surface. The effect is reverse for the cat statue when the metallic strength was reduced. Our method shows similar behavior to the Principled BRDF shaders, which present perceptually subtle effects when tuning the metallic value.

\vspace{1mm}
\noindent \textbf{Albedo.} As the relative strength for the albedo is turned to 1, we observe the albedo of the Buddha statue, fish, and toy go to gray. This is not a trivial in-image desaturation operation as the network maintains the highlights, shadows, and the light tint from the plausible environment map.

\vspace{1mm}
\noindent \textbf{Transparency.} The transparent renditions of the binocular and the chair demonstrate the prior over 3D geometry of the objects, using which it generates the appropriately tinted glass-like appearance and in-paints background objects. With the edit of the banana, we can see the caustics underneath and the preservation of the specular highlights.

\subsection{Baseline Comparisons}
We compare our method, \methodname, to the GAN-based in-image material editing of Subias et al.~\cite{subias2023wild}, Prompt-to-Prompt~\cite{hertz2022prompt} with Null-text Inversion (NTI)\cite{mokady2023null}, and InstructPix2Pix~\cite{brooks2023instructpix2pix} in \Figure{qualitative_results}. Furthermore, we fine-tuned the InstructPix2Pix prompt-based approach with our synthetic dataset. 
Subias et al.'s method results in exaggerated material changes as their objective is perceptual, not physically-based material edits.
Null-text inversion and InstructPix2Pix change the global image information instead of only the object of interest: lighting changes for roughness and albedo edits, or a geometry change for metallicity edit.
When InstructPix2Pix is trained on our dataset with a prompt-only approach, we observe the model exaggerating the intended effect, yielding artifacts on the panda for metallic and the water for the transparency case. 
The model also changes the albedo of the sand when only asked to make the change to the crab. 
Our method faithfully edits only the object of interest, introducing the specular hightlights on the leg of the cat statue, dampening the albedo for the metallic panda, changing the albedo of the crab to gray while retaining the geometry and illumination effects, and turning the dolphin transparent with plausible refractive effects. Specific configurations for each baseline is presented in the supplement.

\input{tables/quantitative_table}

\vspace{1mm}
\noindent \textbf{Quantitative Comparisons.}
Due to the lack of an existing dataset for quantitative analysis, we rendered 10 held-out scenes with unseen 3D objects, materials, and environment maps.
Each scene was rendered from 15 different viewpoints with 10 random scalar values for each attribute. 
We present the average PSNR, SSIM~\cite{wang2004image}, and LPIPS~\cite{zhang2018unreasonable} of edits against GT for prompt-only InstructPix2Pix and \methodname in \Table{quantitative_analysis}.
While the PSNR and SSIM scores are quite close, our model does better in terms of the LPIPS score. 
We also note that prompt-only InstructPix2Pix finetuned on our data does not yield smooth transitions as the relative strength is linearly changed, visible in video results presented in the supplement.
Note that the image reconstruction metrics are not commonly used for evaluation of probabilistic generative models. Samples of the test data and model outputs are presented in the supplement.

\vspace{1mm}
\noindent \textbf{User study.}
To further the comparison between the baseline and our method, we conducted a user study presenting N=14 users with pairs of edited images. For each image pair, the users were asked to choose between the two based on: (1) \textit{Photorealism}, and (2) \textit{``Which edit do you prefer?''}. For both questions, the users were presented with the instruction, i.e. for transparency, the instruction was stated as "the method should be able to output the same object as transparent retaining the tint of the object". Each user was presented with a total of 12 image pairs (3 image results for each of the 4 attributes).

Our method was chosen as the one with more photo-realistic edits (\textbf{69.6\%} vs.\ 30.4\%) and was strongly preferred overall (\textbf{70.2\%} vs.\ 29.8\%). This is likely due to the apparent exaggeration exhibited by InstructPix2Pix trained on our data with prompt-only approach, leading to saturated effects making it less photo-realistic.

\vspace{1mm}
\noindent \textbf{Smoothness in Parametric Control.}
\input{figures/tex/slider_results}
\input{figures/tex/spatial_localization}

We demonstrate that our model achieves fine grained control of material parameters by linearly varying the strength of a single attribute, as shown in \Figure{slider_results}.
Observe that the model generates plausible specular highlights on the headphone instead of naively interpolating pixel values to the extrema and introduces more dampening of the albedo on the cat to give it a metallic look. For transparency, the model preserves the geometry while refracting the light through the object to produce a transparent look. The instructPix2Pix model trained on our data did not yield such smooth results as the relative strength of the attributes were changed in text format. Please refer to the supplementary for video results.

\subsection{Specializations}
\noindent \textbf{Spatial localization.}
Attending to a specific instance of a class when multiple objects of the same class are present in the image is a difficult task using just language instruction.
We explore the possibility of changing the material attribute of a specific instance of the class by only attributing the scalar value in the segmented region, assuming a known segmentation map from an instance segmentation method such as Segment Anything~\cite{kirillov2023segment}.
Though the network was not trained for this specific task, we find that the network \emph{does} respect the localization of the relative attribute strength, though requires over-driving to values beyond $1$. We observe that mask-based editing works in such cases, i.e. changing the material properties of specific instance of cat and cup, as shown in Figure~\ref{fig:spatial_localization}. 

\input{figures/tex/multi_attribute}
\vspace{1mm}
\noindent \textbf{Multi-attribute changes.}
To enable control over multiple attributes in a single diffusion pass, we train our network on two versions of the dataset to be able to vary albedo ($\albedo$), roughness ($\roughness$), and metallic ($\metallic$). In the \textit{axis-only sampled} version, we keep the context image at the baseline, and vary a single attribute at a time, such that only one of $\{\albedo, \roughness, \metallic\}$ is non-zero for any given training target image. In the \textit{volume sampled} version, $\{\albedo, \roughness, \metallic\}$ are all allowed to be non-zero, effectively sampling the 3D volume of material attributes. In both data-sets, we keep the number of input images the same.

We present the qualitative analysis of the joint control in \Figure{multi_attribute}. We find that the "one attribute at a time" model fails to compose the three attributes, generally showing bias towards one of the attributes. The model trained on the volume of these attributes successfully generalizes, demonstrating excellent ability to edit multiple attributes at once. We find this essential to producing a strong metallic appearance on objects, as the Principled BSDF shader requires a white, non-rough, and highly metallic surface to produce this look. 

\vspace{1mm}
\noindent \textbf{Material editing of NeRFs.} \label{sec:nerf_material_editing}
\input{figures/tex/nerf_results}
We test the efficacy of per-frame editing using our method for two-step material control in neural radiance field (NeRF) reconstruction. We use a selection of scenes from the DTU MVS~\cite{dtu_mvs} dataset and edit them to have reduced albedo or higher specular reflections. We train a NeRF with the vanilla configuration based on~\cite{Rebain2022lolnerf} (complete details in the supplement). 

In the results presented in \Figure{nerf_results}, we observe highly plausible renderings from held-out views showing 3D structure with the intended albedo, roughness, and metallic change. Please refer to the supplement for rendered video of NeRFs trained on edited data.

%% file: figures/tex/qualitative_results.tex
\begin{figure*}[t]
    \centering
    \includegraphics[width=\linewidth]{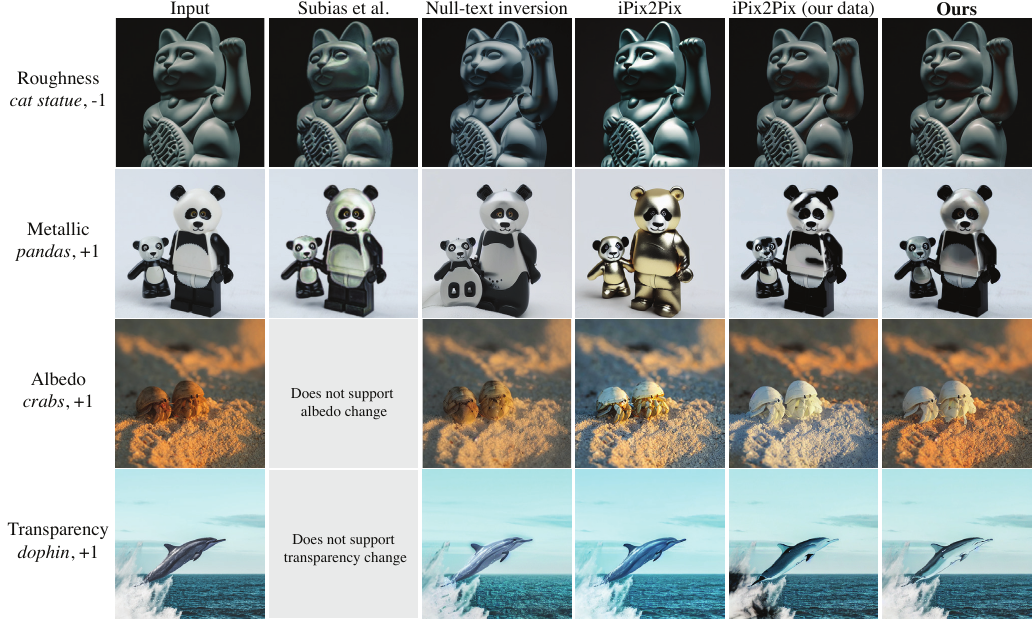}
    \caption{\textbf{Qualitative comparison.} Comparison of \methodname with baseline methods. We increase each of the attributes shown on the left. 
    }
    \label{fig:qualitative_results}
    \vspace{-1em}
\end{figure*}

%% file: tables/quantitative_table.tex
\begin{table}[t]
\centering
\resizebox{\columnwidth}{!}{%
\begin{tabular}{@{}lcccccc@{}}
\toprule
& \multicolumn{3}{c}{InstructPix2Pix w/ our data} & \multicolumn{3}{c}{\textbf{Our Method}} \\ 
& PSNR$\uparrow$ & SSIM$\uparrow$ & LPIPS$\downarrow$ & PSNR$\uparrow$ & SSIM$\uparrow$ & LPIPS$\downarrow$ \\ \midrule
Roughness     & 30.9 & 0.89 & 0.13 & 31.5 & 0.90 & 0.09 \\
Metallic      & 31.0 & 0.89 & 0.10 & 31.1 & 0.89 & 0.09 \\
Albedo        & 26.9 & 0.88 & 0.14 & 27.2 & 0.88 & 0.10 \\
Transparency  & 26.9 & 0.85 & 0.13 & 27.1 & 0.85 & 0.13 \\ 
\bottomrule
\end{tabular}
}
\caption{\textbf{Quantitative analysis.} Metrics for the prompt-only InstructPix2Pix trained on our data and our proposed method computing the PSNR, SSIM~\cite{wang2004image}, and LPIPS~\cite{zhang2018unreasonable} on a held-out unseen synthetic rendered dataset of 10 scenes with 15 cameras.
}
\label{tab:quantitative_analysis}
\vspace{-1em}
\end{table}

%% file: figures/tex/slider_results.tex
\begin{figure*}[ht]
    \vspace{-1em}
    \centering
    \includegraphics[width=\linewidth]{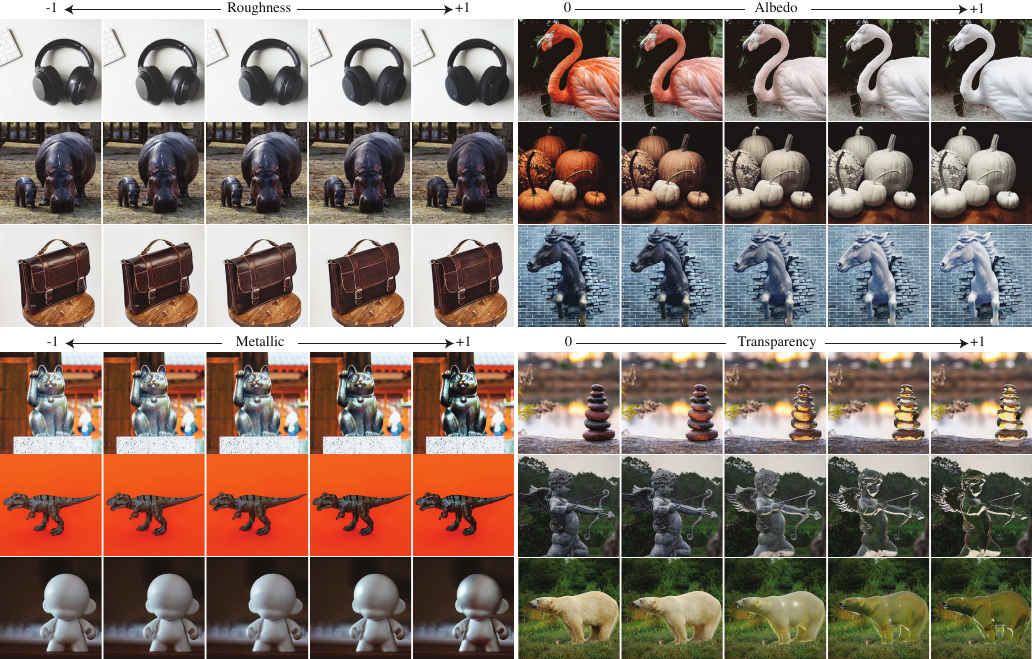}
    \caption{\textbf{Slider results.} \methodname produces edits smoothly with attribute strength. We note that the outputs for linear change in the input relative strength in InstructPix2Pix prompt-only trained on our data results in non-smooth transitions. Refer to the supplement videos.}
    \label{fig:slider_results}
\end{figure*}

%% file: figures/tex/spatial_localization.tex
\begin{figure}[ht]
    \centering
    \includegraphics[]{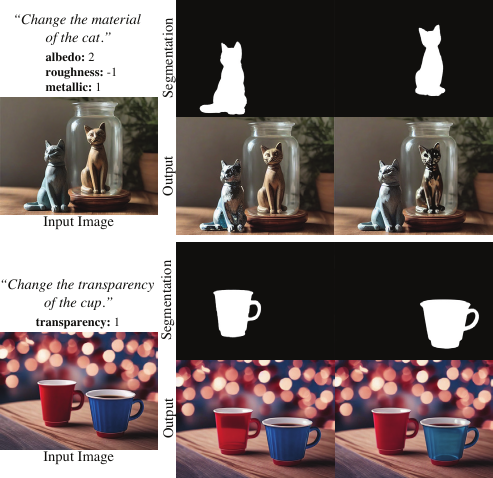}
    \caption{\textbf{Spatial Localization.} Edit results (\textbf{bottom}) when the scalar strength input is masked by the shown segmentation (\textbf{top}). The image is only altered in the segmented region, becoming either shiny (\textit{cat}), or semi-transparent (\textit{cup}).}
    \vspace{-1em}
    \label{fig:spatial_localization}
\end{figure}

%% file: figures/tex/multi_attribute.tex
\begin{figure}[t]
    \vspace{-1em}
    \centering
    \includegraphics[width=\linewidth]{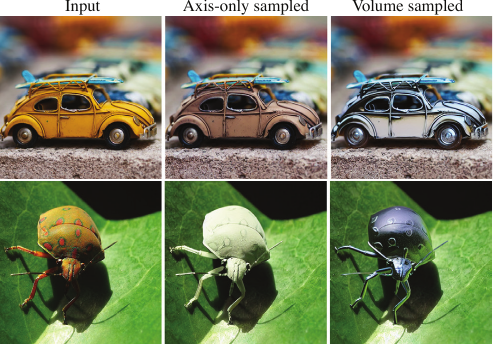}
    \caption{\textbf{Multi-Attribute Editing.} Comparison between an ``axis-only sampled'' model trained on images where only one of $\{\albedo,\roughness,\metallic\}$ is $\ne0$, vs.\ a ``volume sampled'' one where all $\{\albedo,\roughness,\metallic\}$ may be $\ne0$. We show edits with $(\albedo,\roughness,\metallic) = (1,-1,1)$.
    The former tends to edit only a single attribute, while the latter successfully achieves the desired ``silver'' appearance.
    }
    \label{fig:multi_attribute}
    \vspace{-1em}
\end{figure}

%% file: figures/tex/nerf_results.tex
\begin{figure}[t]
    \vspace{-1em}
    \centering

    \includegraphics[width=\linewidth]{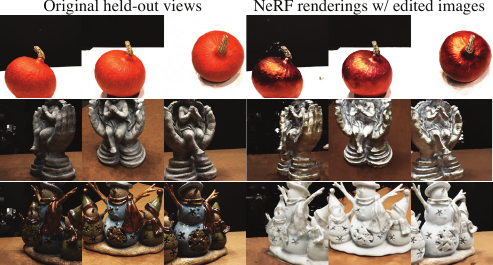}
    \caption{\textbf{NeRF Results. Left:} Original test views from DTU. \textbf{Right:} We edit training views of each scene, train a NeRF, then render held-out test views. The respective edits $(\albedo,\roughness,\metallic)$ from top to bottom are: \textit{scan30}: $(0, -0.5, 0.5)$, \textit{scan118}: $(1, -1, 1)$ and \textit{scan69}: $(1, 1, 0)$.}
    \label{fig:nerf_results}
    \vspace{-1em}
\end{figure}

%% file: sections/5_discussion.tex
\section{Discussion}
\input{figures/tex/limitations}

Our model generalizes to editing fine-grained material properties in real images, despite being trained solely on synthetic data. We believe that our method could extend to a wide range of material alterations achievable with a shader. However, our approach does have limitations, such as producing minimal perceptual changes for roughness and metallic attributes, and occasionally yielding physically unrealistic transparency, as illustrated in \Figure{limitations}. The model lacks a complete 3D world model and is unable inpaint to maintain physical consistency as seen in the candy-cane example. As is typical with generative models, our method generates plausible interpretations that are true to the given instructions, but it does not necessarily replicate the exact outcomes of a traditional graphics renderer.

%% file: figures/tex/limitations.tex
\begin{figure}[t]
    \centering
    \includegraphics[width=\linewidth]{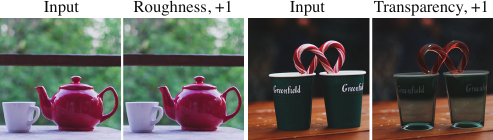}
    \caption{\textbf{Limitations.} \methodname sometimes fails to achieve the desired result. \textbf{Left:} A shiny surface remains on the teapot after a roughness edit. \textbf{Right:} The stem of a candy-cane is omitted.
    }
    \label{fig:limitations}
    \vspace{-1em}
\end{figure}

%% file: sections/6_conclusion.tex
\section{Conclusion}
We present a method that allows precise in-image control of material properties, utilizing the advanced generative capabilities of text-to-image models. Our approach shows that even though the model is trained on synthetic data, it effectively edits real images, achieving seamless transitions as the relative strength of the desired attribute is varied. Beyond image editing, we demonstrate the applicability to NeRF allowing for editable materials in NeRFs. We believe that our work can further impact downstream applications and allow for improved control over low-level properties of objects.